%% file: main.tex
\documentclass[conference]{IEEEtran}
\IEEEoverridecommandlockouts
\usepackage{times}

\usepackage{color}
\usepackage[numbers]{natbib}
\usepackage{multicol}
\usepackage[bookmarks=true]{hyperref}
\usepackage{graphicx}
\usepackage{amsmath}
\usepackage{amssymb}
\usepackage{balance}
\usepackage{multirow}

\usepackage{booktabs}

\begin{document}

\makeatletter

\apptocmd{\@maketitle}{
    \vspace{0cm}~~~~~~~~~~~~~~~~~~~~~~~~~~~~~~~~~~~~~~~~~~~~~~~~~~~~~~~~~~~~~~~~~~~~~~~~~~\insertfig
}{}{}

\makeatother
\newcommand{\insertfig}{%
  \setcounter{figure}{0} 
  \refstepcounter{figure}
  \makebox[0pt]{\includegraphics[width=\linewidth]{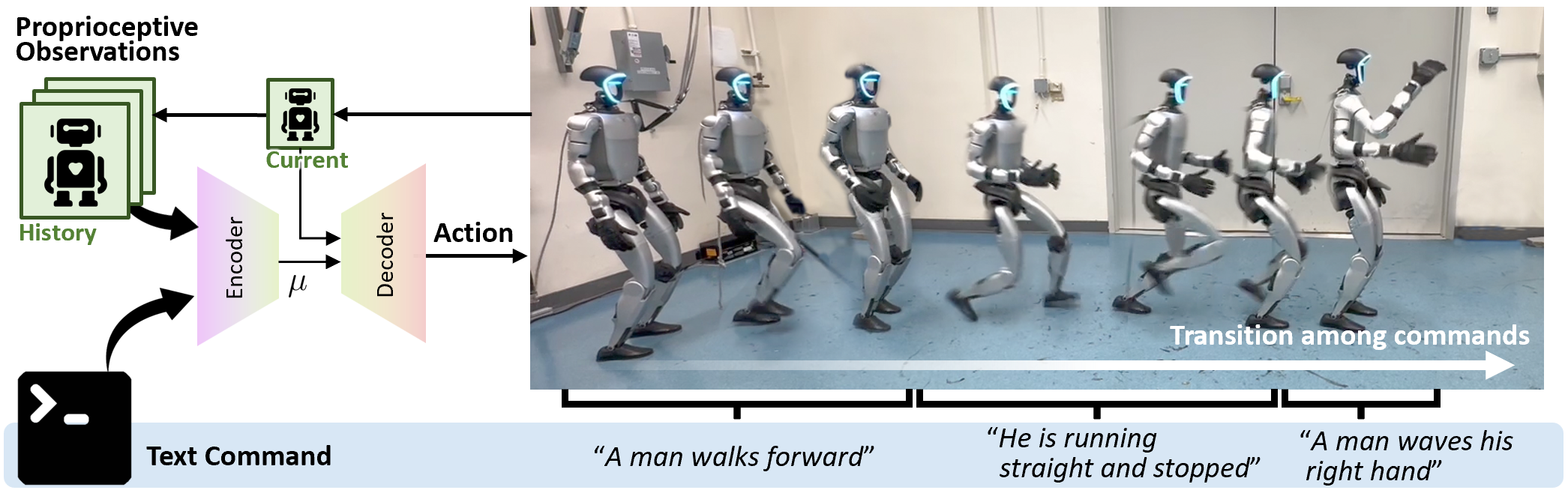}} \\
  \footnotesize{Fig.~\thefigure:\quad We propose a language-directed humanoid whole-body control framework that translates natural language commands into continuous robot actions through a Conditional Variational Autoencoder (CVAE). The structured latent space brought by the CVAE enables smooth transitions between diverse and agile behaviors, as shown in the sequence where the robot seamlessly transitions from walking to running, concluding with a hand-waving motion prompted by the corresponding text commands. See more experiments at \href{https://youtu.be/9AN0GulqWwc}{https://youtu.be/9AN0GulqWwc}}
  \label{fig:cover}
}

\title{LangWBC: Language-directed Humanoid 
Whole-Body Control via End-to-end Learning}


\author{
    Yiyang Shao \quad
    Xiaoyu Huang \quad
    Bike Zhang \quad
    Qiayuan Liao \quad
    Yuman Gao \\
    Yufeng Chi \quad
    Zhongyu Li \quad
    Sophia Shao \quad
    Koushil Sreenath 
    \\
    University of California, Berkeley
}

\maketitle

\begin{abstract}
General-purpose humanoid robots are expected to interact intuitively with humans, enabling seamless integration into daily life. Natural language provides the most accessible medium for this purpose. However, translating language into humanoid whole-body motion remains a significant challenge, primarily due to the gap between linguistic understanding and physical actions. In this work, we present an end-to-end, language-directed policy for real-world humanoid whole-body control. Our approach combines reinforcement learning with policy distillation, allowing a single neural network to interpret language commands and execute corresponding physical actions directly. To enhance motion diversity and compositionality, we incorporate a Conditional Variational Autoencoder (CVAE) structure. The resulting policy achieves agile and versatile whole-body behaviors conditioned on language inputs, with smooth transitions between various motions, enabling adaptation to linguistic variations and the emergence of novel motions. We validate the efficacy and generalizability of our method through extensive simulations and real-world experiments, demonstrating robust whole-body control. Please see our website at \href{https://LangWBC.github.io}{LangWBC.github.io} for more information. 

\end{abstract}

\IEEEpeerreviewmaketitle

\input{section/intro}
\input{section/relatedwork}

\input{section/method}

\input{section/exp}
\input{section/limitation}
\input{section/conclusion}
\input{section/acknowledgement}

\bibliographystyle{plainnat}
\bibliography{references}

\clearpage
\appendix
\input{section/appendix}

\end{document}

%% file: section/intro.tex
\section{Introduction}

Humanoid robots hold immense potential for integration into human environments due to their anthropomorphic design, particularly in areas such as healthcare, personal assistance, and interactive services. For such robots to be truly effective, especially for users with limited technical proficiency such as the elderly, intuitive interaction modalities are essential. Natural language stands out as the most accessible and natural medium for human-robot communication, enabling users to convey complex instructions effortlessly.  

However, translating natural language commands into dynamic, agile, and robust whole-body motions for humanoid robots remains a significant challenge. The fundamental challenge stems from two interconnected aspects: First, as a motion generation problem, the system needs to produce diverse and generalizable movements that accurately reflect the intent of varied natural language commands. Second, as a real-world control problem, it must ensure these generated motions are physically executable while maintaining balance and stability under environmental uncertainties and disturbances. These two aspects are tightly coupled -- generated motions must be physically realizable, while the control strategy must be flexible enough to accommodate the diversity of language-commanded behaviors. 

While prior works on language-directed real-world humanoid control have shown success by decoupling the problem into kinematic motion generation and whole-body tracking control \cite{serifi2024robot, he2024omnih2o, mao2024learning}, this hierarchical approach has key limitations. The generated motions are often physically implausible -- e.g., lower-body floating in the air or upper-body exceeding stability margins -- forcing the tracking policy to trade off between accurately tracking these motions and maintaining balance. Moreover, these methods are restricted to fixed-duration motions, limiting their ability to handle disturbances or ensure smooth transitions between motions.

In this work, we introduce LangWBC, a framework that addresses these dual challenges through a single end-to-end model, eliminating the inherent conflicts between motion generation and physical feasibility. This approach enables humanoid robots to execute agile and diverse whole-body motions from natural language commands with flexible duration, smoothly transition between sequential actions, and synthesize diverse and novel motions through latent space interpolation.

LangWBC uses a two-stage training process. First, a teacher policy is trained via reinforcement learning to track retargeted motion capture (MoCap) data, acquiring a rich repertoire of dynamic and physically plausible behaviors. Then, a student policy based on a Conditional Variational Autoencoder (CVAE) \cite{sohn2015learning} is trained via behavior cloning to learn the mapping from natural language commands and proprioceptive history \emph{directly} to control actions, forming a joint distribution of language and actions within a structured and unified latent space.

We demonstrate the capabilities of LangWBC through extensive simulations and real-world experiments. The robot successfully executes agile motions, including running and quickly turning around, as well as expressive motions like waving and clapping. It also exhibits robustness to disturbances, such as recovering from kicks while executing textual commands. Furthermore, our framework enables smooth transitions between motion clips and generates novel motions through interpolation, demonstrating generalization beyond the training data.

Our key contributions are summarized as follows:
\begin{itemize}
    \item We propose a novel framework that maps natural language commands \emph{directly} to whole-body robot actions in a closed-loop control setup, achieving agile and robust performance suitable for real-world deployment.
    \item Our method enables the generation of diverse motions, smooth transitions, and adaptability to a wide range of textual inputs, including the synthesis of novel behaviors through latent space interpolation using the CVAE architecture.
    \item We validate our approach extensively on a physical humanoid robot, demonstrating its practical applicability, robustness to disturbances, and ability to execute complex whole-body motions from natural language commands.
\end{itemize}

%% file: section/relatedwork.tex
\section{Related Work}

This work explores the intersection of learning-based humanoid whole-body control and generative action modeling. Here, we review prior research in both fields.

\subsection{Learning-based Humanoid Whole-body Control}
Learning-based controllers have demonstrated the ability to perform complex whole-body control for humanoid robots. In physics-based animation, robots have learned various dynamic tasks~\cite{peng2018deepmimic, peng2021amp, tessler2023calm,juravsky2024superpadl}, object interactions~\cite{zhang2023simulation, luo2024grasping, gao2024coohoi}, and even full-body motions from the AMASS dataset~\cite{luo2023perpetual, tessler2024maskedmimic, huang2025diffuse}.

However, transferring these controllers to real-world hardware faces challenges due to the sim-to-real gap. As a result, prior work has largely focused on specialized controllers for a limited set of agile motions on bipedal robots, such as walking~\cite{duan2024learning, liao2024berkeley, radosavovic2024real, gu2024advancing, long2024learning}, jumping~\cite{li2021reinforcement, zhang2024wococo}, and running~\cite{li2024reinforcement, tang2024humanmimic}. More recent efforts aim to scale up the range of feasible humanoid motions. For example, \cite{cheng2024expressive, ji2024exbody2, dugar2024learning} incorporate dozens to a few hundred motions, while \cite{he2024learning, he2024omnih2o, fu2024humanplus, he2024hover} focus on a curated subset of the AMASS dataset. Unlike policies trained for specific skills, these approaches treat motion generation as a tracking problem: the policy learns to track kinematic trajectories but does not inherently compose them into downstream tasks. Instead, an additional high-level planner is required to execute specific motions at test time. In contrast, our work directly learns a text-conditioned motion generation policy, enabling both robust sim-to-real transfer and the ability to generate and execute diverse motions as a downstream task within a single framework.

\subsection{Generative Action Modeling}
Prior approaches to model actions with a generative model can be broadly categorized into two approaches: hierarchical kinematics-based tracking and end-to-end action generation.

\subsubsection{Hierarchical Kinematics-based Tracking}
A common approach is to use a hierarchical framework, where a high-level generative model produces diverse kinematic motions conditioned on inputs such as text~\cite{tevet2023human, shi2024interactive}, keyframes~\cite{cohan2024flexible}, obstacles~\cite{karunratanakul2023guided}, etc., while a low-level tracking controller learns to follow these trajectories. Most prior works adopt this scheme. For example, OmniH2O~\cite{he2024omnih2o} uses a pre-trained fixed-length MDM model~\cite{tevet2023human} for text-conditioned motion generation, followed by a tracking controller. HumanPlus~\cite{fu2024humanplus} uses behavior cloning to learn a high-level policy from human tele-operation to output target poses for downstream tasks. Exbody2~\cite{ji2024exbody2} separately trains a CVAE to generate kinematic motions autoregressively, but lacks text conditioning.

While hierarchical methods have proven effective, they require complex frameworks and often suffer from artifacts or physically infeasible motions in generated trajectories, such as floating bodies, foot sliding, and penetration. To address these challenges, Robot MDM~\cite{serifi2024robot} incorporates a learned Q-function to refine motion generation and enhance feasibility. However, this adds additional training overhead. 

\subsubsection{End-to-end Action Generation}
An alternative is to model control actions directly using a generative approach, eliminating the gap between kinematics generation and tracking. While this method offers advantages in continuity and coherence, it remains under-explored due to its complexity, particularly in high-dimensional dynamic control.

\begin{figure*}[t]
    \centering
    \includegraphics[width=\linewidth]{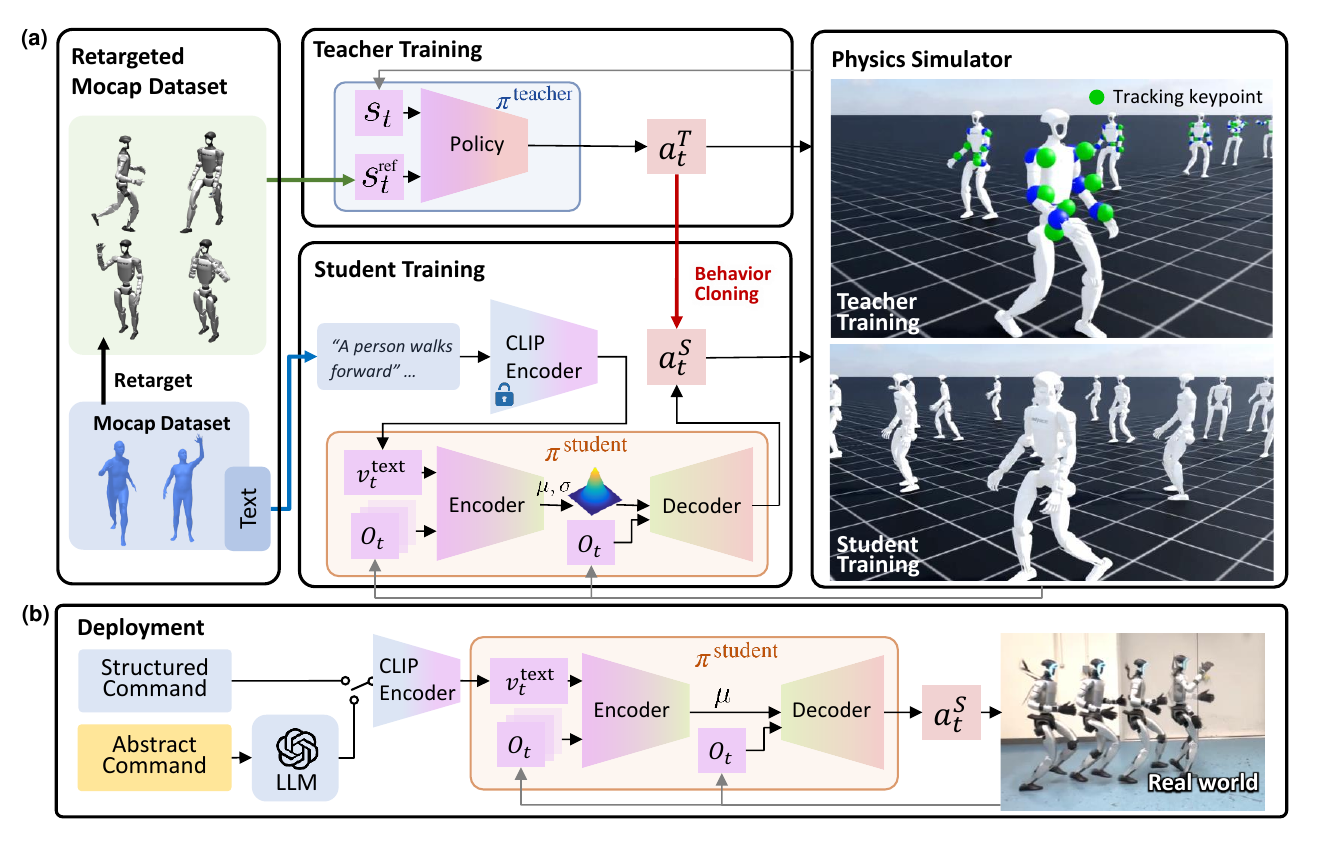}
    \caption{\textbf{The Overview of the Training Framework.} The training process includes a motion-tracking teacher training phase and a language-directed student training phase. We first retarget the MoCap dataset and train a teacher policy via reinforcement learning. Then, a student policy, leveraging a CVAE architecture, jointly models high-level linguistic instructions and low-level physical actions of the teacher policy in a unified latent space. During deployment, we use the student policy for zero-shot sim-to-real transfer on hardware, demonstrating diverse behaviors.  }
    \label{fig:method}
\end{figure*}

In robotic manipulation, diffusion-based policies~\cite{chi2023diffusion, wangequivariant} have demonstrated this approach for quasi-static manipulators. For legged robots, DiffuseLoco~\cite{huang2024diffuseloco} extends it to dynamic control but is limited to state-based commands and quadrupedal robots. A more recent work, UH-1~\cite{mao2024learning}, introduces text-to-action generation but only supports open-loop control, making it less robust to real-world disturbances. In this work, we present a fully functional text-conditioned end-to-end generative controller for humanoid robots. Our model not only enables robust real-world deployment but also generates novel, unseen motions while generalizing to similar text commands.

%% file: section/method.tex
\section{Methods}

In this section, we present LangWBC, an end-to-end framework that jointly models high-level linguistic instructions and low-level physical actions, enabling robots to execute complex whole-body motions directly from language commands.

We begin by training a language-agnostic teacher policy to learn and track a diverse set of human motions. A CVAE student policy is then used to align these physically-plausible motions with language inputs, forming a unified latent space that captures the joint distribution of language and actions. This latent space facilitates generalization, smooth interpolation, and seamless behavior transitions. Simultaneously, by training the CVAE with behavior cloning, we transfer the privileged teacher policy to the student policy that operates solely on proprioceptive inputs, enabling zero-shot sim-to-real transfer using onboard sensors without additional training.

An overview of this framework is illustrated in Fig.~\ref{fig:method}. 

\subsection{Motion-Tracking Teacher Policy}
The teacher policy is designed solely as a motion-tracking policy to track complex human motions without language understanding. The training process of the teacher policy involves two stages, motion retargeting and motion tracking. 

\subsubsection{Motion Retargeting}

To ensure the MoCap trajectories are kinematically feasible for the teacher policy to track, we perform motion retargeting by applying inverse kinematics (IK) based on the Levenberg–Marquardt (LM) algorithm~\cite{marquardt1963algorithm}. We formulate the retargeting as a nonlinear least squares optimization problem that minimizes the position and orientation errors between the robot and MoCap keypoints, while incorporating smoothness constraints to ensure natural transitions between frames. The optimization is solved using the LM algorithm with joint limit constraints, yielding kinematically feasible motions that closely match the original MoCap data. The detailed formulation and implementation are provided in Appendix \ref{appendix:retargeting}.

\subsubsection{Motion Tracking}
The primary objective of the teacher policy is to accurately track the retargeted MoCap trajectories without language information. Therefore, we employ a simple neural network architecture consisting of a multi-layer perceptron (MLP) with layer sizes of 512, 256, and 128 units.

The teacher policy can be formulated as:
\begin{equation}
a_t^{\text{T}} = \pi^{\text{teacher}}(s_t, s_t^{ref}),
\end{equation}
where $s_t\in\mathbb{R}^{175}$ represents robot states, including both proprioceptive states and privileged information (friction, mass, external perturbations, and motor properties) available only in simulation \cite{lee2020learning}, and $s_t^{ref}\in\mathbb R^{141}$ is the reference motion, specifically, the future five-frame keypoint positions in body frame and reference joint positions from the retargeted motion. 
The inclusion of privileged information (in $s_t$) enhances the policy's ability to master complex dynamic skills by providing additional context about the environment and the robot's physical properties. The definition of each specific input state can be found in Appendix \ref{appendix:teacher_policy}. The action output $a^{\text{T}}_t\in\mathbb{R}^{27}$ corresponds to the desired joint positions for the low-level PD controllers. We apply domain randomization for the teacher policy, with details provided in Appendix \ref{appx:domain-rand}. 

Since MoCap datasets contain highly agile motions that are difficult to track in the early stages of training, including the entire datasets often leads to high gradient variance and slow convergence. To improve training efficiency, we design a motion curriculum that gradually increases motion complexity, allowing the policy to adapt progressively to more challenging motions.

We categorize the motions into two levels of difficulty:

a) Easy motions: static or quasi-static movements, typically characterized by low-speed motions.

b) Hard motions: agile motions that require more dynamic whole-body coordination, including actions such as sudden turns or rapid running. 

Training begins with easy motions, and gradually we add hard motions as tracking performance improves. With this curriculum, the teacher policy learns a wide range of physical skills required for executing diverse motions. 

The teacher policy is trained using Proximal Policy Optimization (PPO) \cite{schulman2017proximal} to minimize the discrepancy between the robot's movements and the reference motions. To encourage symmetry in the learned policy, we also incorporate symmetry-based data augmentation and an additional symmetry loss. Specifically, for each state-action pair $(s_t, a^{\text{T}}_t)$, we generate its mirrored counterpart $(s_t^m, a_t^{\text{T}, m})$ through left-right reflection. The augmented training objective is formulated as
\begin{equation}
\mathcal{L}_{\text{teacher}} = \mathcal{L}_{\text{PPO}} + \lambda_{\text{sym}} \mathcal{L}_{\text{sym}},
\end{equation}
where $\lambda_{\text{sym}}$ is a weighting coefficient, and $\mathcal{L}_{\text{sym}}$ encourages consistent policy outputs for mirrored states, i.e., 
\begin{equation}
\mathcal{L}_{\text{sym}} = \mathbb{E}_{s_t \sim \mathcal{D}} \left[|\vert\pi^{\text{teacher}}(s_t) - \mathcal{M}(\pi^{\text{teacher}}(s_t^m))|\vert^2\right].
\end{equation}
Here, $\mathcal{M}(\cdot)$ denotes the mirroring operation for actions. This symmetry constraint helps the policy learn more balanced and natural movements while reducing the sample complexity of training. The tracking reward formulation is summarized in Table ~\ref{tab:reward_functions}. The teacher policy runs at $50$ Hz.

\begin{table}[t]
\centering
\caption{Reward Function Components for Teacher Policy}
\label{tab:reward_functions}
\begin{tabular}{ccc}
\toprule
\textbf{Term} & \textbf{Expression}& \textbf{Weight} \\
\midrule
Z linear vel penalty & $|\vert{v_z^{\text{root}}}|\vert^2$ & $-0.2$ \\
XY angular vel penalty & $|\vert{\omega_{xy}^{\text{root}}}|\vert^2$ & $-0.05$ \\
Joint torque penalty & $\sum \tau_i^2$ & $-2\!\times\!10^{-6}$ \\
Joint acc penalty & $\sum \ddot{\theta}_i^2$ & $-1\!\times\!10^{-7}$ \\
Joint action rate penalty & $\sum (\Delta a^{\text{T}}_i)^2$ & $-0.05$ \\
Energy cost & $\sum |\vert \tau_i \cdot \dot{\theta}_i|\vert^2$ & $-1\!\times\!10^{-6}$ \\
Termination penalty & $\mathbb{I}_{\text{terminated}}$ & $-200$ \\
Joint limit penalty & $\mathbb{I}_{\theta_i \notin [\theta_{\min}, \theta_{\max}]}$ & $-1$ \\
Orientation penalty & $|\vert{{g}_{zy}^{\text{proj}}}|\vert^2$ & $-10.0$ \\
Feet slide penalty & $\mathbb{I}(F_{\text{feet}} > 100\text{N}) \cdot \sum |\vert{v_{xy}^{\text{feet}}}|\vert^2$ & $-0.1$ \\
Hip joint deviation & $\sum |\theta_{\text{hip}} - \theta_{\text{default}}|$ & $-0.03$ \\
Leg joint deviation & $\sum |\theta_{\text{leg}} - \theta_{\text{default}}|$ & $-0.01$ \\
Keypoint tracking & $\exp\left(-\frac{|\vert{{p}_{\text{key}} - {p}_{\text{ref}}}|\vert^2}{2}\right)$ & $1.0$ \\
Joint tracking & $\exp\left(-\frac{|\vert{\boldsymbol{\theta} - \boldsymbol{\theta}_{\text{ref}}}|\vert^2}{4}\right)$ & $1.0$ \\
Single stance reward & $\mathbb{I}_{\substack{\Delta h^{\text{feet}}_{\text{ref}} > 0.05 \\ t_{\text{stance}} \in [0.1,0.5]}} \cdot t_{\text{stance}}$ & $1.5$ \\
\bottomrule
\end{tabular}

\vspace{0.2cm}
\footnotesize
Notations: $\mathbb{I}_{(\cdot)}$ denotes indicator function, $|\vert{\cdot}|\vert$ is Euclidean norm, \\
$\tau$: joint torque, $\theta$: joint angle, $v$: velocity, $\omega$: angular velocity, \\
$F_{\text{feet}}$: ground reaction force, $t_{\text{stance}}$: single-foot stance duration.
\end{table}

\subsection{Language-Directed Student Policy}

To enable the robot to interpret and act on natural language commands, we design a CVAE-based student policy that encodes textual instructions and physical actions into a unified latent space, using only language inputs and proprioceptive readings.

The input of the student policy consists of two parts:
\begin{enumerate}
\item \textbf{Text Caption Embedding}: We utilize the CLIP text encoder~\cite{radford2021learning} to convert the input natural language command $c_t^\text{text}$ into a fixed-length embedding vector
\begin{equation}
v_t^{\text{text}} = f_{\text{CLIP}}(c_t^\text{text}) \in \mathbb{R}^{512}.
\end{equation}
This embedding captures the semantic meaning of the text command.
\item \textbf{History of Proprioceptive Observations}: Instead of providing the full privileged state used in the teacher policy, we provide only proprioceptive observations $o_t \in \mathbb{R}^{90}$ to the student policy, which encapsulate joint positions, joint velocities, base linear velocities, base angular velocities, and projected gravity. We input a sequence of historical observations and actions, sampled at 10 Hz over a 2-second window, yielding a 20-step trajectory of input-output pairs.
\end{enumerate}

The encoder processes the concatenated textual and observational inputs to produce the parameters of a latent Gaussian distribution, outputting a mean vector $\mu \in \mathbb{R}^{128}$ and a diagonal covariance matrix represented by $\sigma \in \mathbb{R}^{128}$. This architecture models the conditional distribution of robot motions given text commands through the latent space, where the text embedding serves as a conditioning signal that shapes the latent distribution. During training, we sample the latent vector $z$ using the standard reparameterization trick:
\begin{align}
    \mu_t, \sigma_t &= \pi^{\text{student}}_{\text{enc.}}(o_t, ..., o_{t-20}, v_t^{\text{text}}),\\
    z_t &= \mu_t + \sigma_t \odot \epsilon_t, \quad \epsilon_t \sim \mathcal{N}(0, I),\\
    a^{\text{S}}_t &= \pi^{\text{student}}_{\text{dec.}}(z_t, o_t),
\end{align}
where $\odot$ denotes element-wise multiplication, $\pi^{\text{student}}_{\text{enc.}}$ is the student encoder, and $\pi^{\text{student}}_{\text{dec.}}$ denotes the decoder. This reparameterization allows gradients to flow through the sampling process. The decoder then takes the sampled latent vector $z_t$ along with the latest state observation to output the action. We use an MLP with layer sizes of 2048, 1024, and 512 units for the encoder, and an MLP with layer sizes of 512, 1024, and 2048 units for the decoder. During inference, we simply use the mean $\mu_t$ of the encoded distribution as the latent vector, eliminating the sampling step to ensure deterministic behavior. The student policy is applied with the same domain randomization as the teacher. 

We employ the Dataset Aggregation (DAgger) algorithm~\cite{ross2011reduction} to train the student policy from the teacher policy with language labels. The training objective follows the variational lower bound 
\begin{equation}
\label{eqn:vae}
\mathcal{L}_{\text{student}} = \|a^{\text{T}}_t - a^{\text{S}}_t\|_2^2 + \lambda_{\text{KL}} \, D_{\text{KL}}(q_{\phi}(z_t|o_{t-20:t}, v_t^{\text{text}}) \| p(z_t)),
\end{equation}
where $D_{\text{KL}}$ is the KL-Divergence operator, and $\lambda_{\text{KL}}$ balances reconstruction quality in behavior cloning with the structural regularization of the latent space.

The training process consists of five steps:

\begin{enumerate}
\item \textbf{Data Collection}: We simulate 1,024 parallel environments. At each time step, the student is given the language command and its history observation.
\item \textbf{Teacher Action Query}: For each state encountered by the student, the corresponding optimal action is obtained by querying the teacher policy.
\item \textbf{Experience Buffer Construction}: We insert the collected student's observations and the teacher's actions to a buffer of 1024 $\times$ 512 ($~500,000$) state-action pairs.
\item \textbf{Loss Computation}: In the early stage of training, the student policy results in large accumulated errors that push the teacher policy out of its training distribution. To mitigate this, instead of tracking absolute positions, the student policy tracks displacement relative to past positions. Let \({p}_t\) be the robot's root position at time $t$, with \(\Delta p_t = {p}_{t} - {p}_{t - \Delta t}\) representing its displacement over interval \(\Delta t\), and \(\Delta {p}_{\text{ref}, t} = {p}_{\text{ref}, t} - {p}_{\text{ref}, t - \Delta t}\) denoting the reference displacement. The robot's tracking objective then becomes minimizing the error between its own displacement and the reference displacement
\begin{equation}
    \min \|\Delta p_t - \Delta {p}_{\text{ref}, t}\|^2.
\end{equation}
This mitigates deviations from the reference motion and preserves the quality of teacher demonstrations.

\item \textbf{Policy Update}: We update the student with the loss in \eqref{eqn:vae}. We use a batch size of 1024 $\times$ 64 and a learning rate of $1\times10^{-5}$, with one epoch per iteration. We then use the student's actions to step the environment. 
\end{enumerate}

We repeat the iterative process, where the student progressively learns to replicate the teacher's behavior while understanding both language inputs and its own observation history. The student policy also runs at $50$ Hz.

\begin{figure}[t]
    \centering
    \includegraphics[trim={0 0 0 0},clip,width=\linewidth]{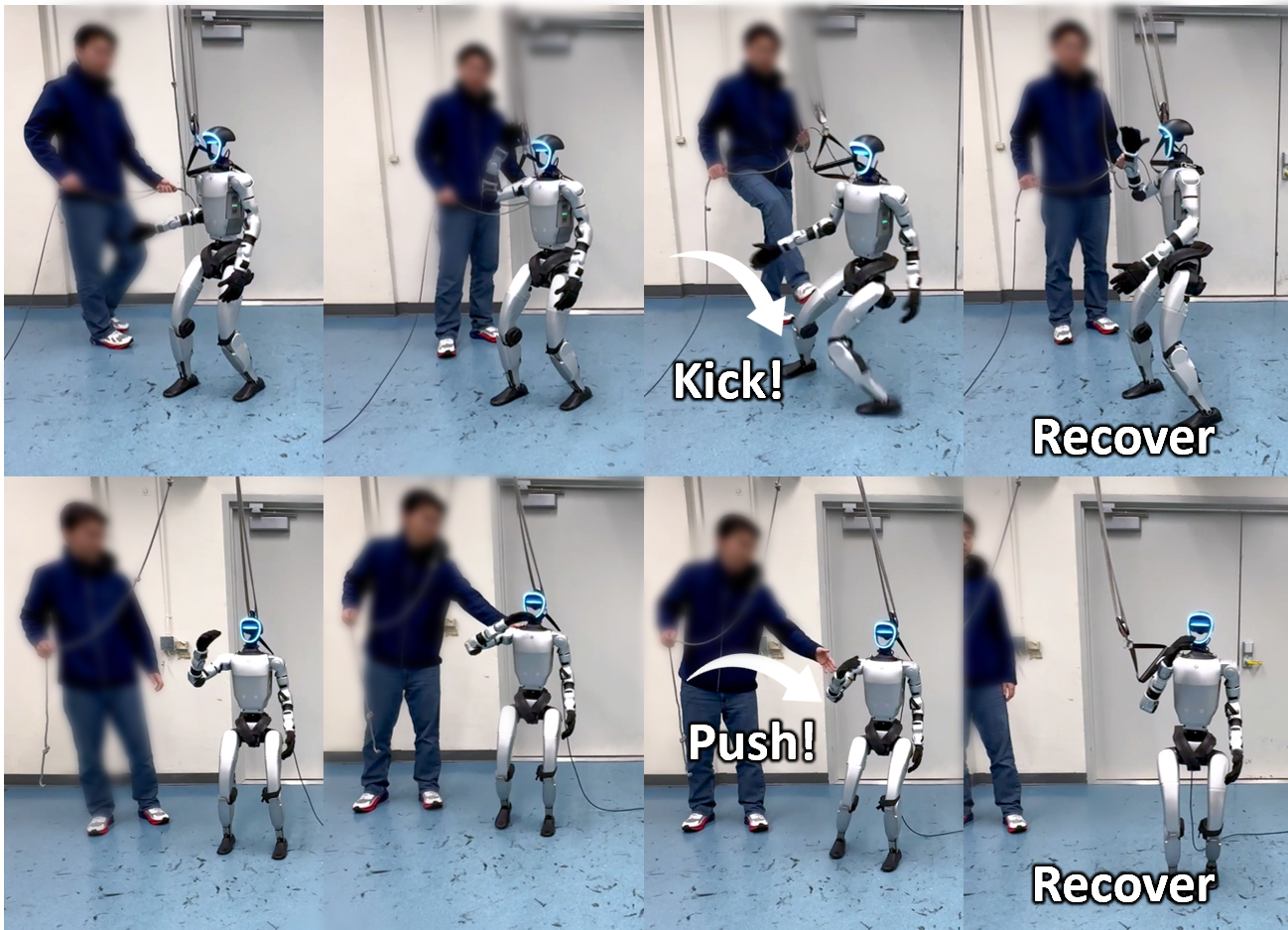}
    \caption{\textbf{Robustness to External Disturbances.} The humanoid robot demonstrates robust stability while executing a hand-waving motion under external perturbations. When subjected to kicks (top row) and pushes (bottom row), the robot maintains balance and continues the commanded motion, showcasing effective disturbance rejection capabilities without interrupting the primary task.}
    \label{fig:exp-kick}
\end{figure}

\begin{figure*}
    \centering
    \includegraphics[width=\linewidth]{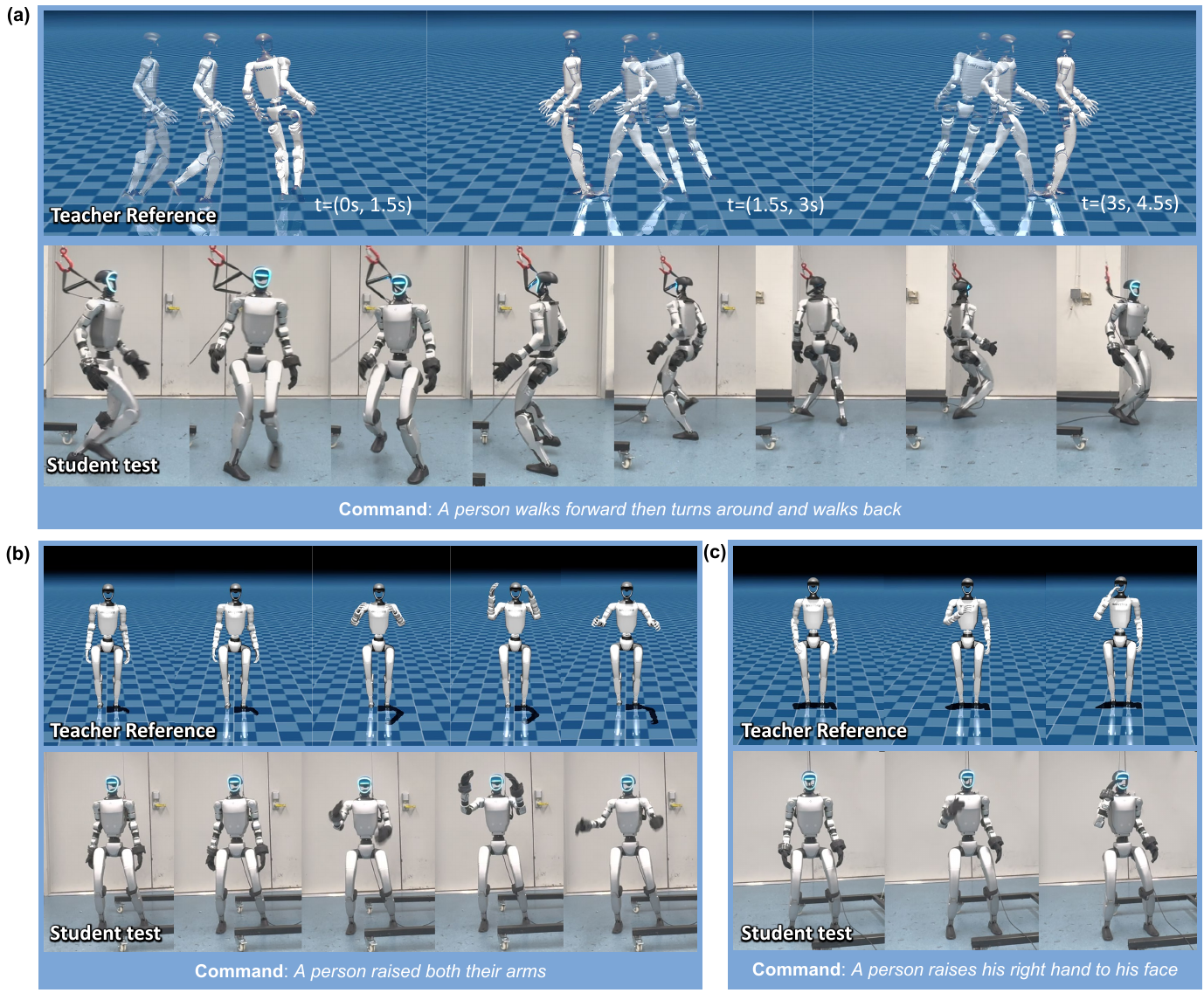}
    \caption{\textbf{Real World Demonstration.} Conditioned on text commands, our framework is able to learn a diverse distribution of whole-body motions in action generation directly, and can be zero-shot deployed on real-world robots. More results are shown in the accompanying video.}
    \label{fig:exp_real_demo}
\end{figure*}

%% file: section/exp.tex
\section{Experiments}

We conduct extensive experiments to evaluate our framework for language-directed humanoid whole-body control with a Unitree G1 humanoid robot. We begin with an overview and demonstrate diverse motions enabled by our approach. We then analyze the learned latent space and its contribution to the policy's generalization to unseen commands, highlight key features such as smooth transitions and latent interpolation, and follow up with an ablation study on core design choices. Finally, we showcase a complex LLM-guided compositional task, illustrating the full capabilities of LangWBC.
We use Isaac Lab~\cite{mittal2023orbit} for training. 

\subsection{Diverse Humanoid Motions}

In order to learn a diverse set of motions, we utilize the HumanML3D dataset~\cite{Guo_2022_CVPR} in training the teacher policy, which provides human MoCap data annotated with textual descriptions. 
For deployment, we use an AMD Ryzen 9 CPU for inference. As shown in Fig.~\ref{fig:exp-kick} and~\ref{fig:exp_real_demo}, the robot successfully executes a diverse range of upper- and lower-body motions in response to natural language commands, including walking in different directions, turning, performing hand gestures, and executing more complex whole-body movements, while remaining robust to external perturbations such as heavy kicks and pushes. For a diverse set of whole-body motions, we provide more results in Appendix~\ref{appx:more-motion}. Through these demonstrations, our framework exhibits zero-shot sim-to-real transfer capabilities, effectively addressing both core challenges through a unified network -- generating diverse, language-aligned motions while maintaining robust control under real-world conditions and disturbances.

\begin{figure}[t]
    \centering
    \includegraphics[trim={0 0 0 0},clip,width=\linewidth]{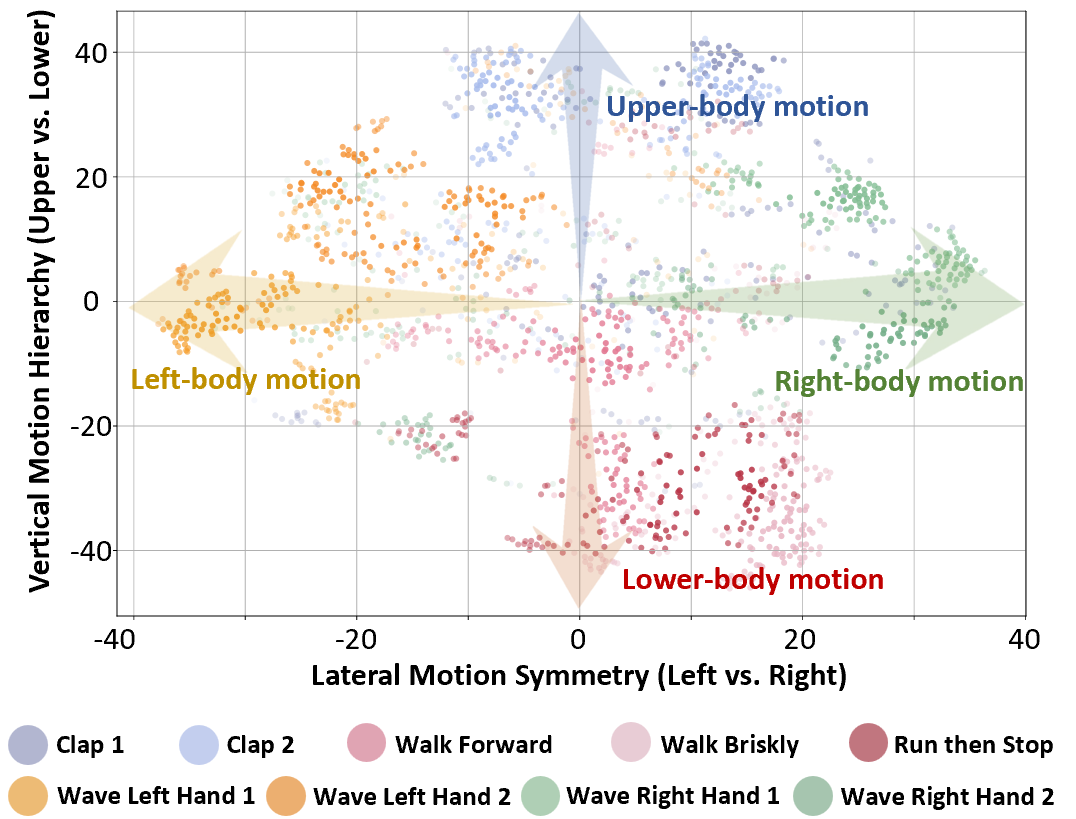}
    \caption{\textbf{t-SNE Analysis of Latent Space. } The plot shows 9 motions from 4 categories of motion, as shown in the legend. We see that similar motions (in the same color band) are closer than dissimilar ones.  The axes suggest an interpretable structure: lateral symmetry (left/right motions mirrored across the y-axis) and vertical hierarchy (upper-body motions cluster at higher y-values, lower-body motions at lower y-values). We observe that all motions share a common region near the origin (0,0), likely representing a typical standing posture.}
    \label{fig:exp-tsne}
\end{figure}

\subsection{Latent Space Analysis}

One key advantage of using a CVAE as the student policy -- rather than a simple MLP network as in previous works~\cite{he2024omnih2o, ji2024exbody2} -- is that it provides a structured latent space. 
This structured latent space aligns language inputs and motion actions to the same latent codes, enabling the model to learn disentangled representations where each latent variable captures both semantic meaning and motion dynamics. As a result, variations in latent space correspond not only to distinct motion styles and transitions but also to meaningful differences in language instructions. This allows for better generalization to unseen commands, smoother motion interpolation, and more coherent transitions between behaviors. 

To verify this property, we apply the t-SNE algorithm to embed the high-dimensional latent codes of various motions into a 2D plane. As shown in Fig.~\ref{fig:exp-tsne}, we plot nine different motions from four categories (walking, raising the left or right hand, and clapping), with each category color-coded.

From the visualization, we see that the latent space has several interpretable features. First, motions in each category form distinct clusters, reflecting clear separation by motion type. Second, there is a striking symmetry: raising the left hand versus the right hand appears mirrored about the center of the latent space, near which more symmetric motions (e.g., walking or clapping with both hands) are located. Third, all motions exist in a common region near the origin, which we interpret as the “standing” latent code -- every motion begins and ends in a standing pose. Overall, this analysis confirms that the CVAE learns a meaningful latent manifold, which we can leverage to perform smooth transitions between motions and generate novel, unseen motions.

\subsection{Generalization to Unseen Texts}
\label{sec:generalization}

Understanding language variations is key to human communication, and essential for flexible communication with robots as well. Since our CVAE's latent space is inherently structured, where semantically similar commands cluster together while remaining distinct from dissimilar ones, we hypothesize it provides robustness to unseen yet contextually similar commands. 

\begin{figure}[t]
    \centering
    \includegraphics[trim={0 0 0 0},clip,width=\linewidth]{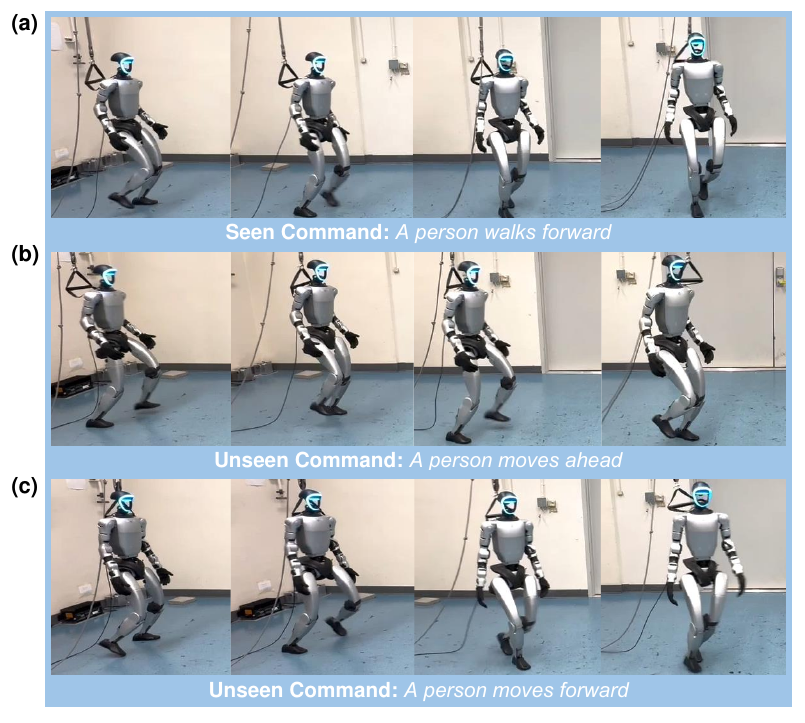}
    \caption{\textbf{Rollouts of Unseen Text Commands.} Our method can generalize to unseen text commands with similar semantical meanings. Of the three, only one command ``A person walks forward" (a) is included in the training dataset. }
    \label{fig:exp-similar}
\end{figure}

To verify this hypothesis, we qualitatively evaluate the policy's response to three semantically similar text commands, as illustrated in Fig.~\ref{fig:exp-similar}. We find the policy performs forward motion in a consistent speed and style despite phrasing differences like “move” vs. “walk,” demonstrating robustness to linguistic variation.

However, this generalization capability may also stem from the CLIP text encoder itself. To isolate the contribution of the CVAE architecture, we compare our approach (\textbf{CLIP+CVAE}) to a baseline that pairs an MLP with a CLIP encoder alone (\textbf{CLIP+MLP}). We evaluate the generated motion quality of both models on a test set of 15 unseen commands spanning three categories:

\begin{enumerate}
    \item \emph{Similar} (e.g., \textit{Walk slowly}),
    \item \emph{Moderately different} (e.g., \textit{Walk into store}),
    \item \emph{Semantically distant} (e.g., \textit{Jump}).
\end{enumerate}

Shown in Table~\ref{tab:unseen_metric},
both variants perform similarly on
commands close to the training data. However, as commands
become more unfamiliar, \textbf{CLIP+CVAE} consistently produces
higher-quality motions. We posit that it is because, while the CLIP
encoder handles minor linguistic variations well, it produces
significantly different encodings for out-of-distribution commands, which the MLP policy struggles to generalize from.
In contrast, the CVAE’s structured joint latent space reduces
the effective distance between novel and seen commands,
mitigating overfitting to the training set and enabling more
plausible motions for out-of-distribution commands.

Summarizing the results, we conclude that our CVAE-based approach generalizes more effectively to language variation, compared to non-structured MLP baselines. This makes it well-suited for integration with large language models (LLMs) in more complex reasoning tasks, as shown in Section~\ref{sec:llm-integration}.

\begin{table}[t]
\centering
\renewcommand{\arraystretch}{1.4}
\setlength{\tabcolsep}{8pt}
\caption{Motion Quality Metric on unseen commands.}
\label{tab:unseen_metric}
\begin{tabular}{lccc}
\hline
\textbf{Model} & \textbf{Similar} & \textbf{Moderate} & \textbf{Different} \\
\hline
CLIP+CVAE     & \textbf{80.92\%} & \textbf{69.58\%} & \textbf{54.62\%} \\
CLIP+MLP     & 80.62\% & 64.20\% & 50.28\% \\
\hline
\end{tabular}

\vspace{0.2cm}
\footnotesize
The Motion Quality Metric is a weighted sum of keypoint and joint errors, normalized by an exponential function to \((0, 1]\) (see Table \ref{tab:reward_functions}). 
\end{table}

\begin{figure}[t]
    \centering
    \includegraphics[trim={0 0 0 0},clip,width=\linewidth]{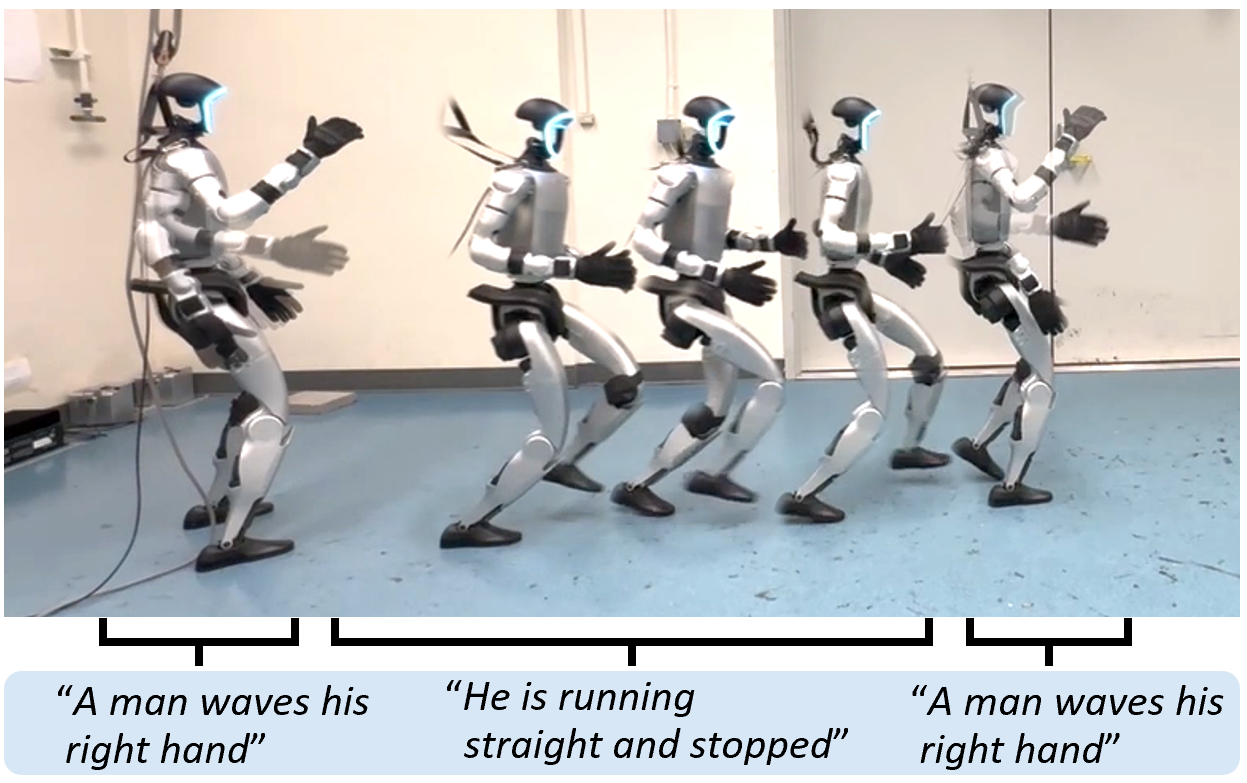}
    \caption{\textbf{Smooth Transitions between Different Text Commands.}
    The humanoid robot seamlessly executes a sequence of actions: waving its right hand, transitioning into running, coming to a stop, and concluding with another hand wave. The policy demonstrates the ability to handle diverse motion transitions within a single execution, without requiring resets between different actions.
    }
    \label{fig:result_trans}
\end{figure}

\begin{figure*}
    \centering
    \includegraphics[width=0.9\linewidth]{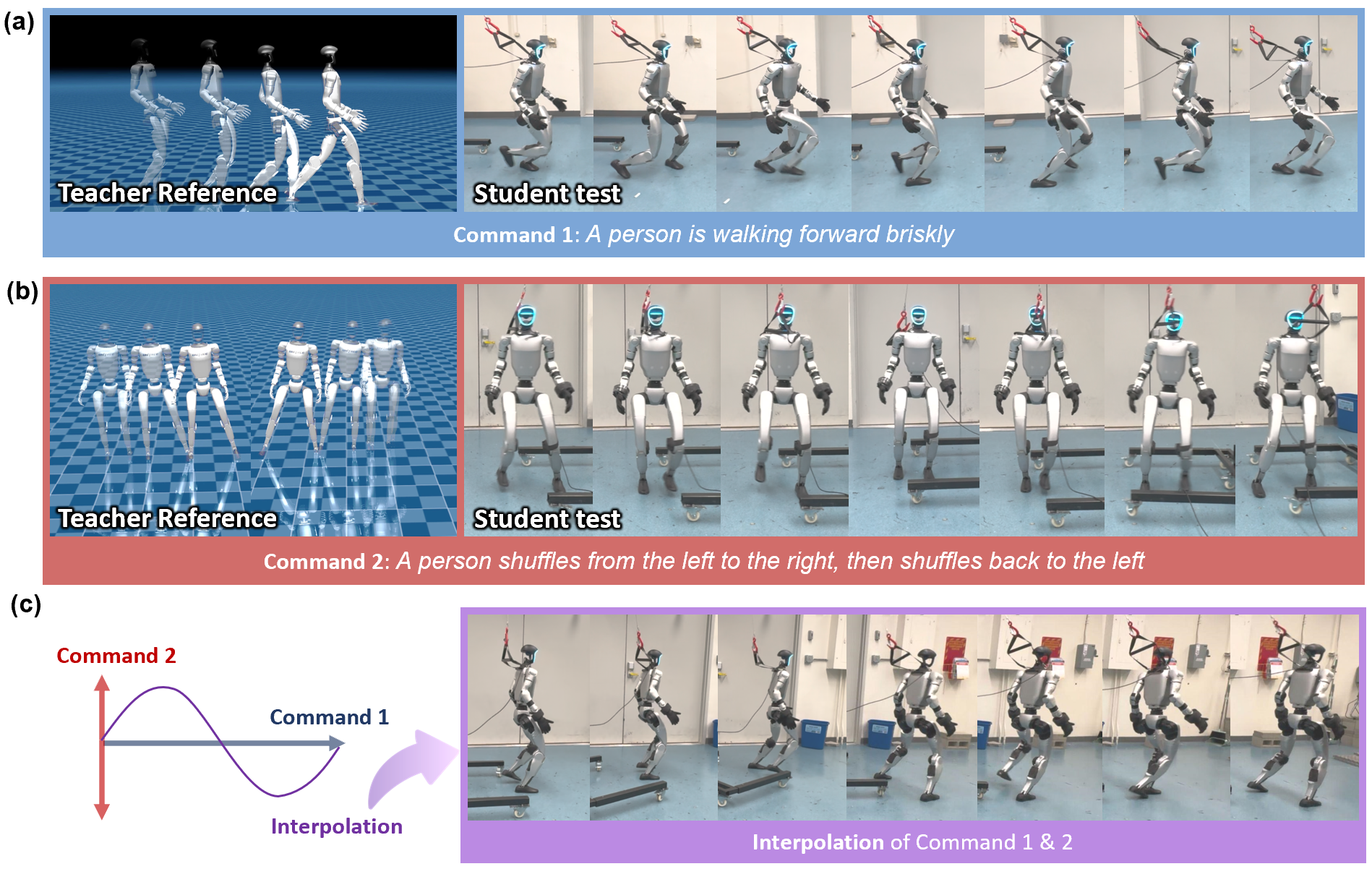}
    \caption{\textbf{Interpolation in the Latent Space.}
    The CVAE gives a structured latent space, enabling the policy to generalize to interpolated command. Here, interpolating between walking (Command 1) and side stepping (Command 2) produces walking to the side, a whole-body motion that does not exist in the training distribution. }
    \label{fig:result_int}
\end{figure*}

\subsection{Smooth Transitions Between Agile Motions}
Furthermore, a major benefit of the structured latent space is the ability to transition smoothly between motions. We demonstrate this capability through examples of agile motion transitions, a challenging task in humanoid control. As shown in Fig.~\ref{fig:cover}, the robot seamlessly transitions from walking to running, then gradually slows to a stop before waving its hand. In another rollout shown in Fig.~\ref{fig:result_trans}, the robot performs an upper-body movement (waving its hand), then begins running, stops, and waves again. Given the high-dimensional dynamics of humanoid motion, achieving smooth and coherent transitions -- such as running, stopping, and switching to limb movements -- within a single policy, without requiring resets, is a significant challenge that has not been demonstrated in prior works. These results underscore the diversity and robustness of our method, as well as its ability to generalize to transition motions that were underrepresented in the training dataset.

Additionally, we observe that the policy can autonomously transition from the end of a motion back to its beginning, enabling seamless looping. As shown in Fig. \ref{fig:exp_real_demo}a, where the reference consists of a single turn, the policy can perform two consecutive turns without interruption. Although the starting and ending poses can be quite different, the policy successfully infers the correct timing for transitioning to ensure a continuous motion.

\subsection{Interpolation in Latent Space}
\label{sec:interpolation}
Since the CVAE creates a smooth and continuous latent space, we find that it also enables the generation of novel motions via a meaningful interpolation between latent codes. This is achieved by encoding the current observation and CLIP-encoded text into latent codes via the CVAE encoder, and then interpolating them in the CVAE latent space and decoding to generate the corresponding action.

To illustrate, we show an example of interpolation between the two distinct commands: \textit{``a man walks forward briskly"} and \textit{``a person shuffles from left to right, then shuffles back to the left"}. As shown in Fig.~\ref{fig:result_int}, the policy generates a novel yet intuitive motion - blending forward walking with lateral shuffling - despite the absence of such behavior in the training data. This interpolated movement emerges naturally as a result of latent space blending, demonstrating the capacity to synthesize new behaviors from learned patterns. Moreover, the robot's movement stays agile and stable, demonstrating the framework's robustness to unseen latent codes. 

Similar to the ablation in Section~\ref{sec:generalization}, since both the CLIP encoder and CVAE architecture have interpolatable latent spaces, this raises a natural question: Is CLIP alone sufficient for interpolating diverse motion commands, or does the CVAE provide essential capabilities? To answer this question, we again isolate the effectiveness of the CVAE latent space by comparing its interpolation performance against a CLIP-only MLP baseline, assessing which latent mixing generates meaningful novel behaviors. 
\begin{itemize}
    \item Interpolating \emph{directly} in the CLIP text–embedding space (\textbf{CLIP Interpolation}).
    \item Interpolating in the \emph{CVAE latent space} learned by the student (\textbf{CVAE Interpolation, Ours}).
\end{itemize}

\begin{figure}[]
  \centering
    \includegraphics[width=0.45\linewidth]{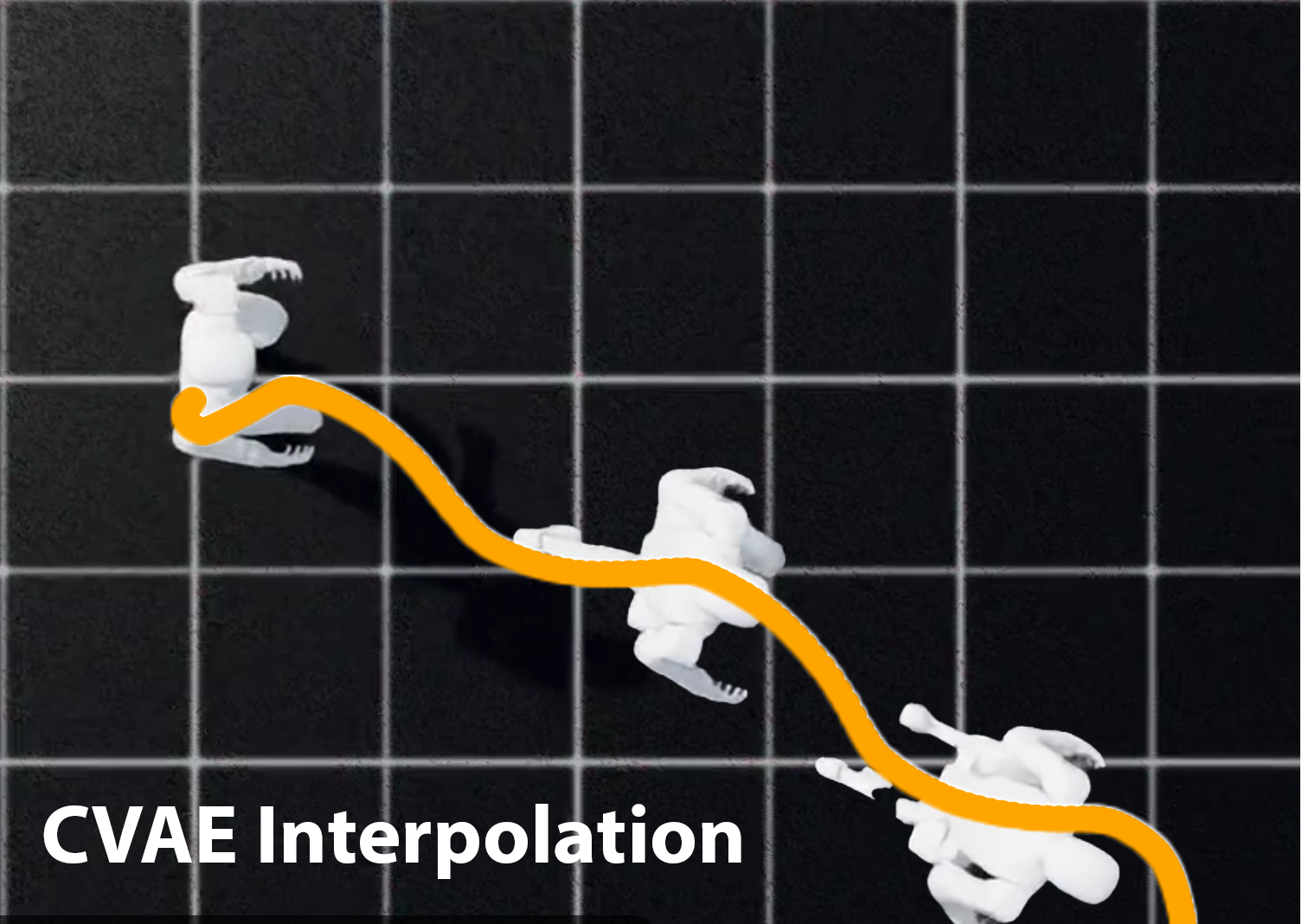}
    \includegraphics[width=0.462\linewidth]{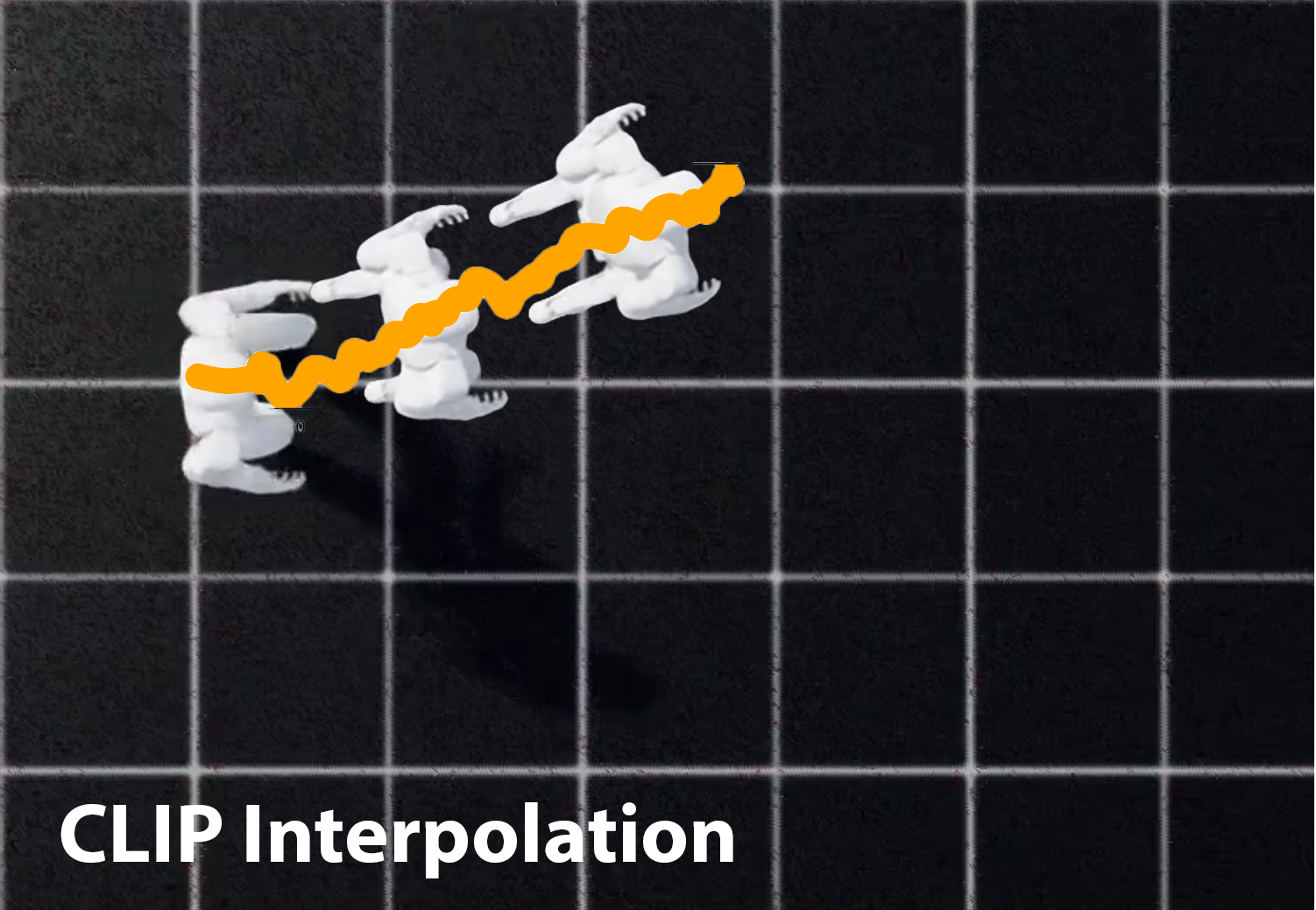}
  \caption{\textbf{Latent Space Interpolation: CLIP+CVAE vs. CLIP Alone} Comparison of motion quality when interpolating between forward and sideways walking. The CLIP+CVAE model (left) produces smooth and coherent diagonal walking, while the CLIP-only baseline (right) results in jittery motions.}
  \label{fig:ablation_cvae}
\end{figure}

As shown in Fig.~\ref{fig:ablation_cvae}, we see that, with the help of the CVAE architecture, the interpolated motion results in a smooth diagonal walk, whereas the baseline with the CLIP encoder alone produces noticeable jitter and difficulty in walking (as seen by the orange CoM
trajectory). This result highlights that the CVAE
induces a smoother and more structured latent space than the
CLIP encoder alone, enabling better generalization to unseen
motions through meaningful motion interpolation.

These results highlight the strength of our CVAE-based architecture in capturing motion space structure, enabling novel motions from dataset diversity. This allows our proposed method to generate flexible, diverse motions and generalize beyond explicit training examples.

\subsection{Ablation Study}

\begin{table}[]
\centering
\renewcommand{\arraystretch}{1.4}
\setlength{\tabcolsep}{8pt}
\caption{Ablation study on the proposed architecture, symmetry loss, and student tracking objectives}
\label{tab:ablation_full}
\begin{tabular}{lcccc}
\hline
\textbf{Metric} & \textbf{CVAE} & \textbf{No‑Symm} & \textbf{No‑Rel} & \textbf{MLP} \\
\hline
Motion Quality $\uparrow$ & \textbf{96.2}\% & 91.6\% & 93.8\% & 91.9\% \\
Stability $\uparrow$      & \textbf{99.10}\% & 98.50\% & 96.60\% & 96.81\% \\
Imitation Loss $\downarrow$ & \textbf{0.085} & 0.107 & 0.098 & 0.091\\
\hline
\end{tabular}

\vspace{0.2cm}
\footnotesize
All metrics are reported after 10k iterations. The Motion Quality metric is defined in Table~\ref{tab:unseen_metric}, the Stability metric indicates success rate over $1000$ steps under perturbations, while the Imitation Loss is the student’s final loss in training.

\end{table}

In the ablation study, we comprehensively evaluate the effectiveness of the three core design choices in our proposed framework: (i) the symmetry loss applied during teacher training, (ii) the relative-tracking objective used for the student, and (iii) the CVAE architecture of the student policy.
Below, we summarize our full framework alongside the ablation baselines used to assess the contribution of each component. We conduct the ablations in simulation. 

\begin{enumerate}
    \item \textbf{LangWBC (CVAE, Ours)}: Our proposed method, in which the teacher is trained with a symmetry loss, and the student leverages a CVAE architecture trained using the relative-tracking objective.
    
    \item \textbf{No Teacher Symmetry (No-Symm)}: A variant of LangWBC in which the symmetry loss is removed from the teacher's training, while all other components remain unchanged.
    
    \item \textbf{No Relative Tracking (No-Rel)}: A variant where the student is trained \emph{without} the relative-tracking objective, while all other components remain unchanged.
    
    \item \textbf{MLP Student (MLP)}: A variant that replaces the student's CVAE architecture with an MLP, while all other components remain unchanged.
    
\end{enumerate}

The quantitative results, summarized in Table~\ref{tab:ablation_full}, indicate that our full framework outperforms all ablation baselines, confirming the contribution of each proposed component to efficient learning and high-quality motion generation. Notably, the CVAE architecture not only enhances generalization, as discussed earlier, but also improves motion tracking accuracy and robustness compared to the MLP baseline. 

\begin{figure}
    \centering
    \includegraphics[trim={0 0 0 0},clip,width=\linewidth]{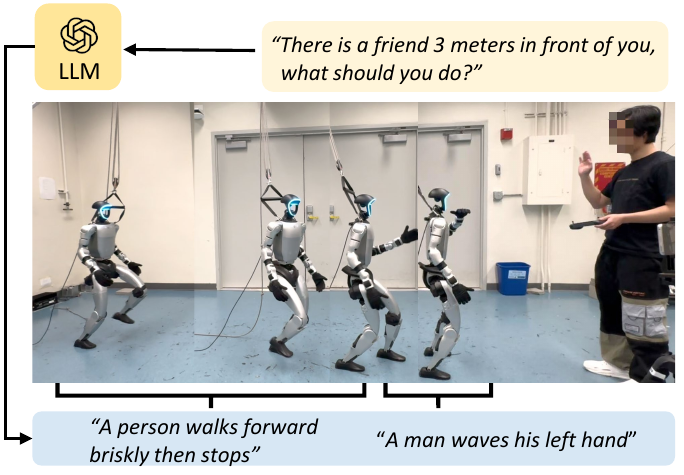}
    \caption{\textbf{LLM-guided Humanoid Motion Sequence.}
    Given the social scenario ``There is a friend 3 meters in front'', the LLM decomposes this high-level instruction into primitive motion commands, which the robot executes by walking forward, stopping, and greeting with a hand wave.
    }
    \label{fig:result_llm}
\end{figure}

\subsection{Integrating LLMs for Complex Tasks}
\label{sec:llm-integration}

To handle abstract instructions and social scenarios, we integrate our framework with a large language model (LLM) as a high-level planner, as shown in Fig.~\ref{fig:method}(b). The LLM translates unstructured natural language inputs into structured command sequences executable by our end-to-end policy. However, LLMs tend to provide commands that are semantically similar to, but not identical to the training data. Thus, the generalization to similar text commands provided by the CVAE becomes critical in this task. 

We prompt the LLM to decompose abstract tasks into motion primitives similar to the training dataset, using the structure ``(Text Command): (Duration in Seconds)." Given the social scenario ``There is a friend 3 meters in front of you, what should you do?", the LLM generates an intuitive sequence of timed commands: 
\begin{itemize}
\item \textit{``A person walks forward briskly then stops: 4.0"}
\item \textit{``A man waves his right hand: 5.0"}
\end{itemize}
During deployment, we run the CLIP encoder at a lower frequency and reuse its embedding across multiple control steps. As shown in Fig.~\ref{fig:result_llm}, the robot successfully performs this socially appropriate sequence, demonstrating our framework’s ability to execute high-level, context-aware instructions. 

%% file: section/limitation.tex
\section{Limitations} 
We have demonstrated a set of humanoid whole-body motions directed by natural language commands. However, the number of language-conditioned motions remains limited to only several dozens of motions due to compute constraints. Expanding the range of language-directed humanoid motions on a larger scale could further validate and enhance our approach. Additionally, the current examples predominantly focus on locomotion-oriented whole-body control due to the absence of a vision module. In future work, we aim to incorporate more whole-body motions that can be leveraged in agile loco-manipulation tasks critical for the deployment of humanoid robots. Due to the limited expressiveness of the variational autoencoder model, we experience some sim-to-real gap which could have been improved via incorporating a more expressive generative model, such as diffusion models, which could capture more subtle diversity in domain randomizations. 

%% file: section/conclusion.tex
\section{Conclusion}
In this work, we presented an end-to-end language-directed humanoid whole-body control framework. It combines learning-based whole-body control with generative action modeling, allowing a single neural network to interpret language commands and execute corresponding physical actions robustly on humanoid hardware in a zero-shot manner. Our framework demonstrates the following characteristics: 

\textbf{Diversity, Generalization, Smooth Transition}: Our framework enables diverse humanoid whole-body motions and composes novel behaviors through latent space interpolation. Leveraging the structured latent space which models the distribution of actions conditioned on language semantics and proprioceptive observations, the policy generates a wide range of motion sequences, including agile behaviors like running, conditioned on language inputs. Furthermore, our method generalizes to contextually similar commands and generates novel motion compositions beyond the training distribution. Moreover, it enables smooth transitions between agile motions, such as from walking to running, with simple language commands, a capability not demonstrated in prior work.

\textbf{End-to-End Learning}: Unlike prior works, our language-directed humanoid whole-body control framework is trained end-to-end, mapping language commands directly to actions without intermediate representations or complex planning modules. By aligning text embeddings with the motion latent space, our approach effectively bridges the gap between language understanding and physical robot skills.

We believe that learning a language-conditioned generative action model for complex humanoid behaviors marks a significant step toward developing a foundation model for humanoid control, paving the way for more intuitive, adaptable, and generalizable deployment of humanoid robots in real-world environments.

%% file: section/acknowledgement.tex
\section{Acknowledgment}
This work was supported in part by The Robotics and AI Institute. We thank Jiaze Cai and Kaixuan Wang for help with hardware experiment. We thank Vishnu Sangli for helpful discussions. K. Sreenath has financial interests in The Robotics and AI Institute. He and the company may benefit from the commercialization of the results of this research.

%% file: section/appendix.tex
\subsection{Teacher Policy Input State}
\label{appendix:teacher_policy}

The input of the teacher policy is defined as follows.

\begin{itemize}
\item \textbf{Base Linear Velocity} (3 dimensions): The robot's current linear velocity in the base frame.
\item \textbf{Base Angular Velocity} (3 dimensions): The robot's current angular velocity in the base frame.
\item \textbf{Projected Gravity} (3 dimensions): The gravity vector projected onto the robot's coordinate frame.
\item \textbf{Actual Keypoint Positions} (7 links $\times$ 3 dimensions = 21): The positions of key body links (e.g., limbs and torso) in the robot's body frame that need to be tracked.
\item \textbf{Joint Positions} (27 dimensions): The current positions of the robot's joints.
\item \textbf{Joint Velocities} (27 dimensions): The current velocities of the robot's joints.
\item \textbf{Last Action} (27 dimensions): The last action executed by the robot.
\item \textbf{Privileged Information} (64 dimensions): Simulation-exclusive information encompasses: 1) physical properties including static friction, dynamic friction, and restitution coefficients (3 dims); 2) joint armature (27 dims); 3) body link mass (28 dims); 4) external force (3 dims); and 5) external torque (3 dims).
\item \textbf{Target Keypoint Positions} (6 frames $\times$ 7 links $\times$ 3 dimensions = 126): Future desired positions of keypoints over six time steps, providing a trajectory for the robot to follow.
\item \textbf{Degree-of-Freedom (DoF) Commands} (15 dimensions): Commands specifying desired joint configurations.
\end{itemize}

\subsection{Motion Retargeting Details}
\label{appendix:retargeting}

We formulate the retargeting problem as a nonlinear least square optimization over the entire motion sequence. Let \(T\) be the number of frames, and \({ q}_t\in \mathbb {R}^n\) be the vector of robot joint angles at frame \(t\). Our objective is to find the set of joint angles that minimize the following cost function:
\begin{equation}
\label{eq:retargeting_objective}
\begin{aligned}
\min_{\{ {q}_t \}} \quad & \sum_{t=1}^T \Bigl( \left\| {x}_{\text{robot}, t}({q}_t) - {x}_{\text{mocap}, t} \right\|^2 \\ 
& + w_{\text{ori}} \left\| \Delta {r}_{t}({q}_t) \right\|^2 \Bigr) + w_{\text{smooth}} \sum_{t=2}^T \left\| {q}_t - {q}_{t-1} \right\|^2, 
\end{aligned}
\end{equation}
subject to:
\begin{equation}
\label{eq:joint_limits}
{q}_{\min} \leq {q}_t \leq {q}_{\max}, \quad \forall \, t = 1, \dots, T.
\end{equation}
To solve this optimization problem, we employ the Levenberg–Marquardt algorithm, which is suitable for solving nonlinear least squares problems in IK. At each iteration, we linearize the residuals around the current estimate \({q}_t^{(k)}\) and compute the update \(\Delta {q}_t = {q}_t^{(k+1)} - {q}_t^{(k)}\) by solving:
\begin{equation}
\label{eq:lm_update}
\left( {J}_k^\top {J}_k + \lambda {I} + w_{\text{smooth}} {S}^\top {S} \right) \Delta {q} = -{J}_k^\top {f}_k.
\end{equation}

Where:
\begin{itemize}
    \item \({x}_{\text{robot}, t}\) denotes the positions of the robot's keypoints at frame \(t\) as a function of the joint angles \({q}_t\).
    
    \item \({x}_{\text{mocap}, t}\) represents the target keypoint positions from the mocap data.
    
    \item \(\Delta {r}_{t}({q}_t)\) represents the orientation error between the robot's end-effectors and the mocap data at frame \(t\), calculated using the Lie-algebra error.
    
    \item \(w_{\text{ori}}\) is the weighting factor for the orientation error term.
    
    \item \(w_{\text{smooth}}\) is the weighting factor for the smoothness term.
    
    \item \({q}_{\min}\) and \({q}_{\max}\) define the joint limit boundaries.
    
    \item \(\Delta {q} = [\Delta {q}_1^\top, \Delta {q}_2^\top, \dots, \Delta {q}_T^\top]^\top\) is the stacked update vector for all frames.
    
    \item \({J}_k\) is the Jacobian matrix of residuals at iteration \(k\), combining position and orientation terms across all frames.
    
    \item \({f}_k\) is the stacked residual vector at iteration \(k\), defined as:
    \begin{equation}
    {f}_k = \begin{bmatrix}
    {x}_{\text{robot},1}({q}_1^{(k)}) - {x}_{\text{mocap},1} \\
    \sqrt{w_{\text{ori}}} \Delta {r}_{1}({q}_1^{(k)}) \\
    \vdots \\
    {x}_{\text{robot},T}({q}_T^{(k)}) - {x}_{\text{mocap},T} \\
    \sqrt{w_{\text{ori}}} \Delta {r}_{T}({q}_T^{(k)}) \\
    \sqrt{w_{\text{smooth}}} ({q}_2^{(k)} - {q}_1^{(k)}) \\
    \sqrt{w_{\text{smooth}}} ({q}_3^{(k)} - {q}_2^{(k)}) \\
    \vdots \\
    \sqrt{w_{\text{smooth}}} ({q}_T^{(k)} - {q}_{T-1}^{(k)})
    \end{bmatrix}
    \end{equation}.
    
    \item \({\lambda}\) is the damping parameter that balances the Gauss-Newton and gradient descent methods, ensuring convergence and stability in the LM algorithm.
    
    \item \({I}\) is the identity matrix.
    
    \item \({S}\) is the sparse matrix representing the smoothness constraints, defined as:
    \begin{equation}
    {S} = \begin{bmatrix}
    -{I} & {I} & & & \\
    & -{I} & {I} & & \\
    & & \ddots & \ddots & \\
    & & & -{I} & {I}
    \end{bmatrix}
    \end{equation}.
\end{itemize}

The term \( w_{\text{smooth}} {S}^\top {S}\) in Eq.~\eqref{eq:lm_update} adds regularization that enforces smooth transitions between consecutive frames, promoting natural motion. By solving this equation, we obtain the updates \(\Delta {q}_t\) for all frames, which are used to update the joint angles:
\begin{equation}
\label{eq:joint_update}
{q}_t^{(k+1)} = {q}_t^{(k)} + \Delta {q}_t.
\end{equation}
After each update, we enforce the joint limits by projecting \({q}_t^{(k+1)}\) onto the feasible set defined in Eq.~\eqref{eq:joint_limits}. 

Iteratively applying the LM algorithm under these constraints yields kinematically feasible motions that closely match the original mocap data, thus providing suitable reference trajectories for the teacher policy.

\begin{figure*}
    \centering
    \includegraphics[width=\linewidth]{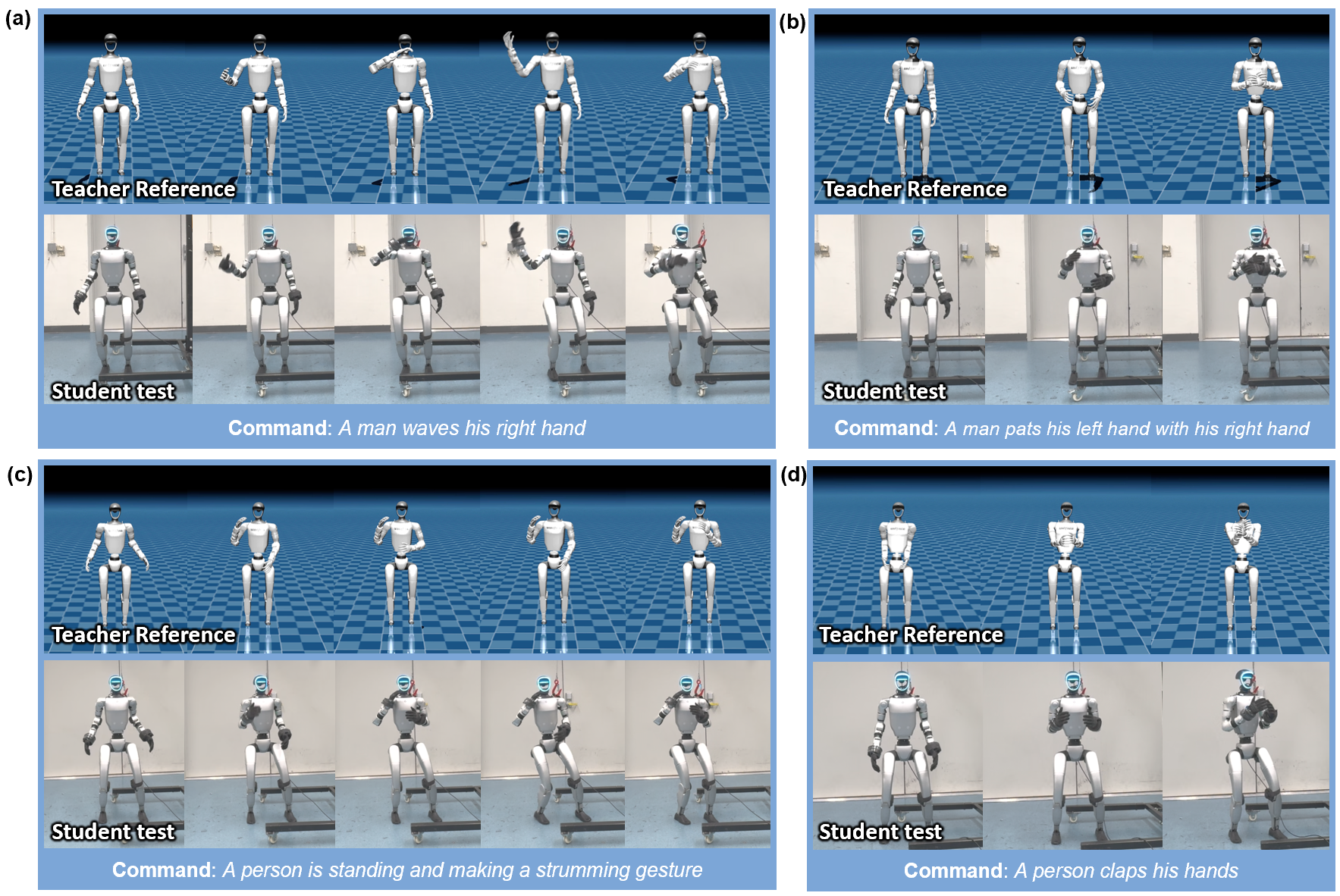}
    \caption{\textbf{Upper-body Motion Examples.} Our framework generates diverse upper-body movements including reaching, lifting, and manipulation tasks. The learned motions can be directly deployed on the real robot.}
    \label{fig:more_exp_1}
\end{figure*}

\begin{figure*}
    \centering
    \includegraphics[width=\linewidth]{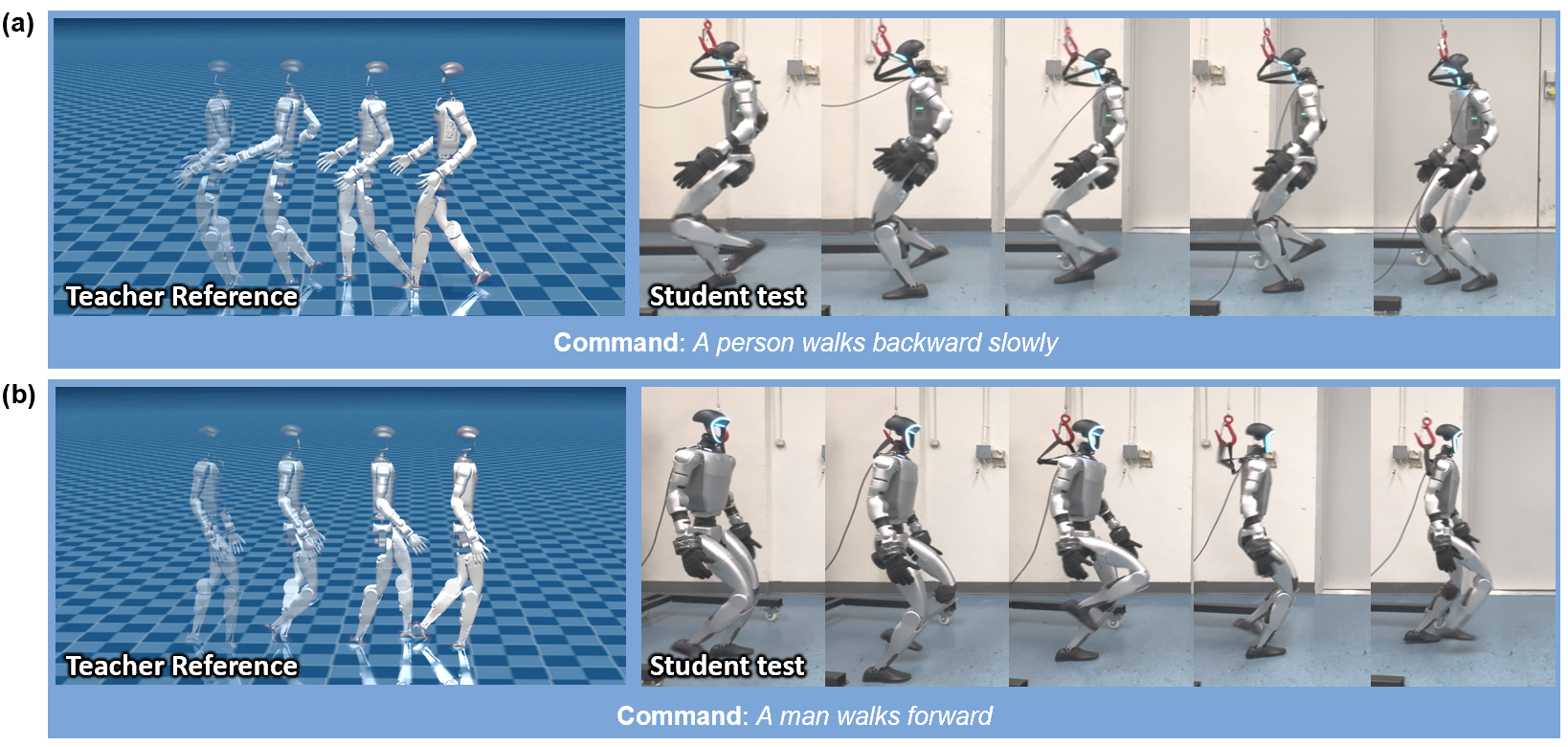}
    \caption{\textbf{Lower-body Motion Examples.} The framework also enables various lower-body movements such as stepping, squatting and balancing. These motions are also successfully transferred to the real robot without additional training.}
    \label{fig:more_exp_2}
\end{figure*}

\subsection{Diverse Whole-body Motions on Real Robot}
\label{appx:more-motion}
We demonstrate our framework's capability to generate and execute diverse whole-body motions on real hardware. As shown in Fig. \ref{fig:more_exp_1} and Fig. \ref{fig:more_exp_2}, our method can learn a rich distribution of both upper-body and lower-body movements, and successfully transfer them to the physical robot with zero-shot. The accompanying video provides more comprehensive demonstrations of these motions.

\subsection{Domain Randomization Configuration}
\label{appx:domain-rand}
Table~\ref{tab:domain_randomization} describes the domain randomization parameters applied to the robot . 
"Reset" parameters are applied at environment reset, "Startup" parameters are applied once during initialization, 
and "Interval" parameters are applied periodically during simulation.
\begin{table}[t]
\centering
\caption{Domain Randomization Parameters}
\label{tab:domain_randomization}
\begin{tabular}{ccc}
\toprule
\textbf{Mode} & \textbf{Component} & \textbf{Range} \\
\midrule
\multirow{2}{*}{Reset} & Initial velocity (linear) & $[-0.5, 0.5]$ m/s \\
 & Initial velocity (angular) & $[-0.5, 0.5]$ rad/s \\
\midrule
\multirow{2}{*}{Reset} & Joint positions & $[0.5, 1.5] \times$ default \\
 & Joint velocities & $[0.0, 0.0]$ rad/s \\
\midrule
\multirow{3}{*}{Startup} & Ankle friction (static) & $[0.2, 0.6]$ \\
 & Ankle friction (dynamic) & $[0.2, 0.6]$ \\
 & Ankle restitution & $[0.0, 0.4]$ \\
\midrule
\multirow{3}{*}{Startup} & Link masses & $[0.9, 1.1] \times$ default \\
 & Base mass modification & $[-1.0, 1.0]$ kg (additive) \\
 & Joint armature & $[0.8, 1.2] \times$ default \\
\midrule
Startup & Joint default positions & $[-0.05, 0.05]$ rad (additive) \\
\midrule
Reset & Base torque & $[-5.0, 5.0]$ Nm \\
\midrule
Interval & Push robot (every 10-15s) & $[-1.0, 1.0]$ m/s (x,y) \\
\bottomrule
\end{tabular}

\end{table}

\balance